# The Dynamics of Intelligence Explosions

Toby Ord*

AI is increasingly being used to help with AI R&D. Under certain conditions this feedback loop might be able to produce an intelligence explosion, with rapidly escalating AI capabilities. I explore the mathematics of the most explosive possibilities, with an eye to understanding what drives the dynamics. I show that singular growth (towards a vertical asymptote) is harder to achieve than would be expected from recent economics-inspired modelling, and that there is an important but neglected class of growth rates that are faster than exponential but don't lead to a vertical asymptote. I draw out the *generation time* (the time to go around the feedback loop) as a neglected parameter that plays a pivotal role in determining the behaviour of any intelligence explosion — one cannot have singular growth unless the generation time rapidly approaches zero.

**Keywords:** recursive self-improvement, RSI, intelligence explosion, explosive growth, finite time singularity, generation time.

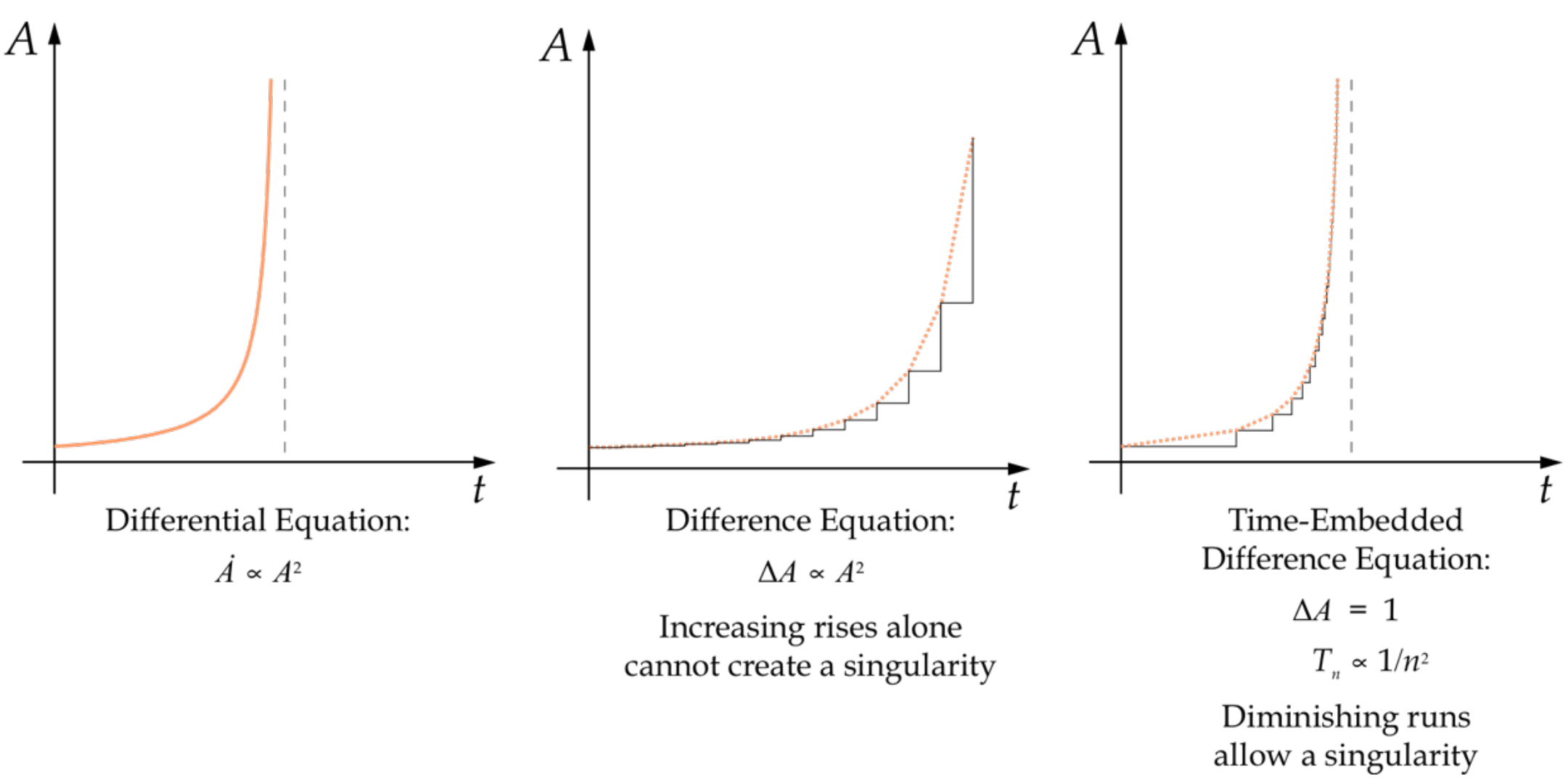

*Preview of Figure 1.* A vertical asymptote requires the gradient (rise over run) to approach infinity within a finite time. Decreasing the run is key. It cannot be achieved with a fixed feedback generation time (centre) no matter how quickly the improvements grow, but can occur when the generation time approaches zero (right) even with a fixed additive improvement.

---

* Affiliation: Oxford Martin AI Governance Initiative, at Oxford University. With thanks to Tom Davidson, Rohin Shah, Loren Fryxell, Fin Moorhouse, Matthew van der Merwe, Will MacAskill, Robert Trager, Simon Biggs, and Shamil Chandaria for helpful discussions.

**The Possibility of an Intelligence Explosion**

AI can be applied to automate many different things. One of those is the R&D that goes into building better AI systems. There have already been minor examples of partially automating this process, such as using AI to: find better optimisation algorithms for training neural networks (Andrychowicz et al. 2016), design a more efficient matrix multiply algorithm (Fawzi et al. 2022), help write the code for new AI systems (Anthropic 2026), and run experiments to iteratively improve AI systems (Karpathy 2026). Leading AI companies are increasingly talking about using an AI system to do more and more of the work of designing and building its successor — something they hope (and fear) could lead to a radical acceleration in the rate of progress in AI (Hassabis et al. 2026).

I. J. Good (1965) introduced the idea of AI systems increasing their own intelligence in what he called an 'intelligence explosion':

> Let an ultraintelligent machine be defined as a machine that can far surpass all the intellectual activities of any man however clever. Since the design of machines is one of these intellectual activities, an ultraintelligent machine could design even better machines; there would then unquestionably be an 'intelligence explosion,' and the intelligence of man would be left far behind...

Implicit in Good's intelligence explosion is the idea that this process of designing a better machine can be iterated. $AI_1$ can create $AI_2$, which can create $AI_3$, and so on. Each system is getting better at intellectual activities — including that of AI R&D — such that each successive system should become ever more intellectually capable.

Following the recent literature, I'll refer to this process as recursive self-improvement, or RSI (Yudkowsky 2001). I use this term very inclusively, to cover situations where humans are still doing almost all of the work through to situations where AI is doing most of the work — or even all of the work. I include cases where the AI is taking its own files as input and improving them (self-improvement), as well as cases where it is building an entirely new AI system. And I include cases where the intelligence level of successive systems 'explodes', as well as cases where it merely improves the speed of progress by some multiple, or where the gains quickly fizzle out.

By accelerating the already rapid progress in AI capabilities, RSI might be extremely dangerous. One reason is that it would likely speed up the risk-creating processes relative to risk-reducing ones (Vaintrob & Cotton-Barratt 2025). For example it may speed up the development of AI capabilities relative to: technical AI safety research, deliberation at the company, and society's ability to understand and respond. RSI could also provide the opportunity for a single unaligned AI to deliberately corrupt all subsequent AIs. And by enabling larger jumps in capabilities per model release, RSI would remove any opportunity society has to learn from ill effects of intermediately powerful AIs. Finally, by widening the capability gap between a leading AI system and a rival that is a few months behind, RSI would increase the likelihood of a winner-take-all dynamic.

These could provide strong reasons against pursuing RSI. This might warrant legislation restricting it, industry best practices restricting it, or internal company policies restricting it. Such restrictions could take many forms including: careful monitoring, retaining meaningful human control, speed limits on capability increase per month, enforced pauses when various capability levels are reached, or bans on certain approaches to RSI.

My focus here, however, is not on risks (nor responses to risks), but on understanding the dynamics of RSI. In particular I aim to improve our understanding of the qualitatively different kinds of explosive growth that could occur, what conditions produce them, and how likely they are to be possible. While it is entirely possible that RSI might lead to interesting and important growth in AI capabilities that fall short of explosive progress, my focus in this paper is on better understanding the most extreme kinds of growth.

I'll show:

- how a common approach to modelling RSI conflates two very different kinds of explosive growth,
- how consideration of the physical duration of the feedback loop (the generation time) is key,
- how a super-exponential period of progress would likely end when this generation time can no longer be reduced,
- and how this makes growth with a vertical asymptote somewhat less likely.

**Modelling RSI through Differential Equations**

A number of recent articles have attempted to mathematically model the possibility of explosive growth from RSI (Aghion et al. 2017, Erdil et al. 2024, Eth & Davidson 2025, Kokotajlo & Lifland 2025). A common approach is to use differential equations connecting the rate at which the AI's capability increases to its current level of capability. This captures the idea that the amount by which a system can improve its capability in the next time period depends on its current level of capability. While differential equations aren't the only way to model RSI, they are a logical choice, being the standard way to model feedback loops in physics, engineering, and economics.

Let's define $A$ to be some measure of the cognitive capability of an AI system (we'll say more about *which* measure later). Adopting the Newtonian notation for differential equations, we use $\dot{A}$ to represent the derivative of $A$ with respect to time. This is a more compact notation for $\frac{dA}{dt}$ — the rate at which $A$ is changing per unit time.

Let's start with what may be the most well-known differential equation of all:[1]

(1) $\dot{A} = kA$

This says that the rate of change of $A$ is in direct proportion to its size. For example, a simple model for the rate of change of a population of animals is that it is in direct proportion to the number of animals, with $k$ set by the difference between the birth rate and the death rate. Solving this differential equation to find how $A$ evolves as a function of time famously gives exponential growth (starting from some initial population, $A_0$):

(2) $A(t) = A_0 e^{kt}$

But what if $\dot{A}$ doesn't vary in direct proportion to $A$? What if there are increasing returns, such that doubling $A$ more than doubles $\dot{A}$? Or decreasing returns, such that doubling $A$ less than doubles $\dot{A}$?

Economists typically allow for this by raising $A$ to some power, $r$.

(3) $\dot{A} = kA^r$

This is a flexible approach, covering many different regimes of growth. When $r = 1$, this is the same equation as before, producing exponential growth. When $r = 0$ there is linear growth. When $r < 0$ growth is sub-linear. When $0 < r < 1$, $A(t)$ grows faster than linear, but slower than exponential (e.g. $r = \frac{1}{2}$ gives quadratic growth). And finally, when $r > 1$, $A(t)$ grows more quickly than an exponential. In particular, it grows as a hyperbola, reaching a vertical asymptote at some finite time $t^*$.

This means that as time increases towards $t^*$, $A(t)$ grows explosively, with every finite level of $A$ being exceeded prior to time $t^*$. In such cases it is said that $A(t)$ has a mathematical *singularity* at $t^*$. This term 'singularity' has been adopted by futurists to refer to various kinds of pivotal moment in the future of technology — often with an almost mystical undertone. But for the purposes of this essay, it simply means that the mathematical model of how $A$ changes with time has a vertical asymptote at some particular future time $t^*$.

Economists often use formulas like (3) in endogenous growth theory. This typically involves a set of differential equations connecting total economic output ($Y$), labour/population ($L$), capital ($K$), and productivity/technology/knowledge/ideas ($A$).

[1] We could also write this as an explicit function of $t$, such as $\dot{A}_t = kA_t$ or $\dot{A}(t) = kA(t)$. But we'll follow the standard shorthand of dropping the explicit reference to $t$ where possible.

For example, Kremer (1993) looked at the very long run history of economic growth and population, finding that the population growth rate and economic growth rate have increased substantially over the last million years, with the growth rate being roughly proportional to the size of the population at that time. He modelled this with a set of differential equations and found that they produced hyperbolic growth of population and productivity.

In such models, $\dot{A}$ is typically expressed as a product of several factors, one of which is $A$ raised to a power. While this power is often restricted to be $\leq 1$, there is usually another factor (such as $L$) which also grows as a function of $A$. Once the full set of differential equations is solved, we sometimes see that $A$ is effectively raised to a power greater than 1, allowing hyperbolic growth and its finite time singularity.

Several recent models of RSI start with semi-endogenous growth theory. This is a theory introduced by Jones (1995) in which explosive growth is harder to achieve due to diminishing returns in the growth of ideas. Its central equation is:

(4) $$\dot{A} = \delta L^{\lambda} A^{1-\beta}$$

Where: $\delta$ is the productivity per researcher; $\lambda$ ($\leq 1$) represents the *stepping-on-toes effect*, where having twice as many researchers at the same time is less than twice as productive due to issues of duplication and coordination; and $\beta$ ($> 0$) represents the *fishing-out effect*, where subsequent ideas get harder to find.

Holding population ($L$) constant, this would prevent even exponential growth in technology ($A$), let alone a singularity. In Jones's original paper he suggests that while the direct effect of $A$ on $\dot{A}$ has diminishing returns, it has also allowed an exponential increase in population, and it is *this* that drives the exponential rise in technology.

Recent work applying this to RSI has often focused more on improving the efficiency of AI than on improving its *intelligence*. For example, Davidson et al. (2026) model a situation where AI is able to perform R&D as well as a human and let $A$ be a measure of its computational efficiency. This sidesteps extremely thorny issues of how to measure intelligence and opens up a clever way of getting hyperbolic growth out of Jones's model.[2] The key is that the total amount of AI labour will be the product of the total compute devoted to RSI ($C$) multiplied by the AI's computational efficiency[3]:

(5) $$L = C\,A$$

We can then plug this back into equation (4) getting:

(6) $$\dot{A} = \delta C^{\lambda} A^{\lambda} A^{1-\beta} \quad = \quad \delta C^{\lambda} A^{1+\lambda-\beta}$$

Now $A$ is being raised to the power of $1 + \lambda - \beta$, which could be greater than 1, so could produce hyperbolic growth. Indeed, if we assume that the growth in compute will be much slower than this self-reinforcing growth in efficiency (so effectively hold $C$ constant), and we simplify the equation by defining $r$ to be $1 + \lambda - \beta$, then we have an equation with exactly the original form of (3). So even in the relatively tame semi-endogenous growth theory, the possibility of a singularity is back on the table.

(It is worth noting that by defining $A$ as a measure of efficiency, this isn't really a model of an *intelligence* explosion at all. It is an *efficiency explosion*. Or more precisely, a *labour explosion*. This makes the model something of a lower bound on what might happen, since at every point AI also has the option of doing R&D to increase its intelligence and presumably the optimal path involves both efficiency and intelligence improvements. Even if this is a useful lower bound, it does mean there remains an important gap of *actually modelling an intelligence explosion*.)

As well as the recent flurry of economics-inspired papers, there is a little-known earlier literature by computer scientists. Figures such as Solomonoff (1985), Kurzweil (2001), and Moravec (1999, 2003) also used sets of differential equations to model the dynamics of an intelligence explosion. One key difference is allowing a greater flexibility of functional forms for the relationships. See Sandberg (2013) for an excellent survey.

**A Note on Singularities**

Before we look into modifying this standard approach, I want to clarify something important about models like these that involve singularities.

---

[2] The same move was made much earlier by Solomonoff (1985) and Moravec (1999, 2003) in their differential equations for intelligence explosions from AI labour speeding up Moore's Law.

[3] Defining labour in this way effectively means we are assuming there is *only* AI labour. While slightly more complicated, it is also possible to model a mixture of human and AI labour, with the AI share increasing over time as it becomes more efficient (Davidson & Houlden 2025).

While the model has an important quantity ($A$) rising to infinity in finite time, I'm not sure whether anyone working in this field believes this will actually happen. Instead, they typically believe that the model will cease to match reality at some point before $t^*$.

One reason for this is that there may be some upper limit to how high $A$ can go — either a conceptual limit or a practical limit. For instance, we know that exponential growth is often a good model of the start of a compounding process, but eventually runs into some limiting factor. When we zoom out, we see that it was really just the beginning of a larger S-curve, plateauing at some finite size. In such cases, we can say that the early part of the curve was approximated very closely by an exponential, but there was really some additional term in the equation (for an effect like overcrowding) which started very small, but eventually came to dominate the longterm behaviour.[4]

This could easily be true for intelligence explosions too. If so, then hyperbolic growth might fit the true trajectory of $A$ very closely until it gets into the vicinity of some upper bound $A^*$, where it starts to fit more and more poorly. Proponents of these models are merely saying that there could be some substantial part of the trajectory of $A$ where the model is a good fit — better than equally simple alternatives such as exponentials (Sandberg 2013). If $A$ only plateaus at some point many orders of magnitude above its starting point, then even if the hyperbola only fits for a small domain of time, it might fit for a very large range of capabilities. If it predicts those much better than alternatives — and *explains* why things are increasing so fast — it would count as a successful model. It's OK for a model to only have a finite domain of applicability.

Just as exponential growth can be a great model for the start of a process, so can hyperbolic growth. Hyperbolic growth will probably cease to be a good model at an earlier time, though not necessarily at a lower height on the graph.

In physics, singularities in the models of particular systems are often seen as useful pointers to where those models must break down and new (hitherto unmodelled) behaviour must begin. We could also adopt that frame here and see the models as pointing to some finite time $t^*$, before which some new unmodelled aspect of the system must take over.

So we'll explore the behaviour of mathematical models of RSI — which sometimes include a singularity — but will remember that something will probably stop the real physical process prior to that point. And we will explicitly return to these limitations in the final section.

---

[4] For example, taking the exponential, $\dot{A} = kA$, and subtracting a quadratic term that has been shrunk down by a large factor, $K$, gives the differential equation for logistic growth, $\dot{A} = kA - kA^2/K$, which has a plateau at height $K$.

## Generalising the Standard Differential Equation for RSI

If we take a closer look at equation (3) ($\dot{A} = kA^r$) we might wonder why it takes this particular form. Taking an arbitrary power of $A$ is certainly a natural and convenient way to create an adjustable differential equation that allows for both sub-exponential and super-exponential growth. But it is one very specific way of doing it, raising the question of whether some of the results about singularities depend on this particular form. We might ask:

- Are there differential equations that produce intermediate patterns of growth lying between exponential and hyperbolic growth?
- What feature of the differential equation is producing the singularity? Is it super-linearity? Being everywhere convex? Something else?
- Do we still get a singularity if we make slight changes to the function (such as adding a constant, or inducing a slight wobble)?

Let's find out, starting with the more general form of the first-order autonomous differential equation:

(7) $\dot{A} = f(A)$

What conditions do we need to impose on $f$ to get super-exponential growth? Does this then always lead to a singularity?

Super-exponential growth means that $A(t)$ eventually overtakes every exponential. Since exponentials have constant relative growth rate ($\frac{\dot{A}}{A} = k$) we require that this relative growth rate ($\frac{f(A)}{A}$) overtakes all constant levels:

(8) $\lim_{A \to \infty} \frac{f(A)}{A} = \infty$

In other words, $f$ must grow super-linearly in $A$. This ensures the right behaviour when $A$ is high enough.

To this we must add a second condition to ensure $A$ can grow high enough:[5]

(9) $f(A) > 0$ at every point in $[A_0, \infty)$

---

[5] As written, this is quite restrictive since it implies $A$ is monotonically increasing. However, that is an artefact of this simple autonomous setup where $\dot{A}$ is just a function of $A$. If you allow it to be a function of $A$ and $t$, or instead look at $\ddot{A}$ as a function of $A$ and $\dot{A}$, then you can also have trajectories that dip for a while before blowing up to a vertical asymptote. For that richer class of differential equations, one would need to define a less restrictive version of this condition.

Let's call this the *positivity condition*. Together, these conditions are sufficient for super-exponential growth.

What is required of $f$ for $A(t)$ to possess a singularity? This comes down to whether it takes a finite time for $A(t)$ to approach an infinite height. We can work out the time at which $A(t)$ reaches a given height $A'$, by starting at $A_0$ and adding up the amount of time needed to accumulate each small gain in height $dA$. The time needed for each gain is just the reciprocal of the slope ($\frac{1}{f(A)}$) times $dA$. So the total time to reach a height $A'$ is just:

(10) $$t(A') = \int_{A_0}^{A'} \frac{1}{f(A)} dA$$

And thus the time it takes to climb all the way from $A_0$ to $\infty$ is:

(11) $$t^* = \int_{A_0}^{\infty} \frac{1}{f(A)} dA$$

When this integral converges to a finite value, it means that there is a finite time by which $A(t)$ goes to infinity — i.e. a singularity. The convergence of this integral is the key condition for $A(t)$ having a singularity, which we shall call the *blow-up condition*. To this we again add the positivity condition — $f(A) > 0$ at every point in $[A_0, \infty)$ — which ensures $A(t)$ is growing (rather than shrinking) and that there is no division by zero.

In the standard differential equations for RSI (where $\dot{A}$ is some power law of $A$) everything that meets the super-linearity condition also meets the blow-up condition: so everything that is super-exponential has a singularity.

But we can now show that this is *not* true in general. Consider:

(12) $$\dot{A} = A \log(A)$$

$A \log(A)$ is super-linear, so $A(t)$ grows super-exponentially. But it doesn't meet the blow-up condition: the integral of $\frac{1}{A \log(A)}$ diverges to infinity, so it takes infinitely long for $A$ to grow infinitely large. Solving the differential equation reveals that $A$ in fact grows doubly exponentially with time:

(13) $$A(t) = A_0^{e^t}$$

This is super-exponential but has no singularity. So if RSI were to obey this equation, there would be a very rapid rise in capabilities, but one of quite a different kind. It would always have more room for improvement, even after an arbitrarily long period of growth.

And this isn't the only function that produces such growth. For instance, consider the infinite family of equations:

- $\dot{A} = A\log(A)$
- $\dot{A} = A\log(A)\ \log(\log(A))$
- $\dot{A} = A\log(A)\ \log(\log(A))\log(\log(\log(A)))$
- …

These all fail to meet the blow-up condition, so do not possess singularities. Instead they grow as a double exponential of time, a triple exponential, a quadruple exponential, and so on. Yet they also come extremely close to the blow-up condition. If you raise the final factor in any of them to a power greater than one (e.g. $A\log(A)^{1+\epsilon}$), they satisfy the blow-up condition and $A(t)$ will have a singularity.

(See the appendix for a table showing the rates of growth corresponding to a wide range of $f(A)$, including whether they produce singularities.)

So in the general case where $\dot{A} = f(A)$, there is a narrow zone of functions for $f(A)$ that fit between the exponential growth given by $\dot{A} = kA$ and the growth towards a finite time singularity given by $\dot{A} = kA^{1+\epsilon}$. We thus can't assume that all super-exponential growth has a singularity.

We might wonder whether it is realistically possible to end up in this zone. Economists call exponential growth a 'knife-edge solution' to $\dot{A} = kA^r$, suggesting that it is vanishingly unlikely for $r$ to take on the value of *exactly* 1. If so, this new zone I'm pointing to may look like it is on the edge of the edge of the knife. Or it would if we were taking this functional form and supposing a random value of $r$. But the assumption of the functional form is then doing most of the work. If we instead suppose an unknown functional form for $f$, with a non-zero chance of it being $A\log(A)$, the landing zone looks a little larger.

I think this functional form is plausible when the $A$ and the $\log(A)$ are coming from different places. For example, suppose $A$ is a measure of the total number of AI researchers doing RSI, and suppose that the contribution of each one towards increasing $A$ is proportional to the log of their total population (due to weak spillovers from each person's research, or weak economies of scale). Then you get a combined effect where $\dot{A} = kA\log(A)$.[6]

---

[6] I subsequently discovered that Kurzweil (Kurzweil, Vinge, and Moravec 2003) suggested this $\dot{A} = kA\log(A)$ form was more plausible than those with $A$ raised to powers greater than one, since it avoided the vertical asymptote. Instead it 'merely' produces doubly exponential growth which he thought was already visible in the long run data (Kurzweil 2001). And he justified $A\log A$ via the same decomposition I gave above. Despite being closely associated with the idea of a 'technological singularity' he found dynamics leading towards a mathematical singularity unlikely — 'It is hard to explain how we could get infinite knowledge, or infinite information processing, from a finite world'.

That said, it is still a narrow zone and any kind of fishing-out or stepping-on-toes effect could easily knock things out of the zone. Overall, I think this possibility would be just something of an interesting footnote were it not for new considerations we'll see in the next section, which substantially widen this zone.

Finally, it is important to note that in this general case of $\dot{A} = f(A)$, one can't treat all growth with a singularity as hyperbolic. That was only true when making the modelling assumption that $f(A)$ has the form of a power law. Some functions (such as $\dot{A} = A\log(A)^2$) produce trajectories that grow more quickly than any hyperbola — though they are simultaneously less 'explosive' in the sense that the growth isn't as packed into the final moment. Others (such as $\dot{A} = e^A$) produce trajectories that grow more slowly than any hyperbola, but are more explosive since they still reach infinity in finite time, despite lagging behind until the last moment. To reflect this wider class of functions, I'll use the more inclusive term *singular growth* to refer to all cases when $A(t)$ possesses a singularity (whether hyperbolic or not).

We can now answer our earlier questions:

- There are differential equations that produce intermediate patterns of growth lying between exponential and singular growth, such as $\dot{A} = A\log(A)$, which gives doubly exponential growth.
- $f(A)$ being super-linear in the limit produces the super-exponentiality, but more is needed to create a singularity. Being everywhere convex is neither necessary nor sufficient — singularities are not produced by satisfying a local property like convexity, but by satisfying a global property where deficient growth in $f(A)$ somewhere can be made up by faster growth elsewhere (so long as it is always positive). The relevant global property is the blow-up condition: that the integral of the reciprocal of $f(A)$ converges.
- Small changes to $f$, such as small additive or multiplicative constants or small random deviations from its path won't usually change these behaviours, unless they make $f(A) \leq 0$ somewhere along the trajectory.

**Feedback loops & discrete timesteps**

The idea of an intelligence explosion relies on some kind of feedback loop, where each AI system builds a more intelligent successor, which is even better at AI R&D. There are many forms this feedback loop could take. For example:

- AI could develop new tools and processes for making faster chips.
- AI could create better chip designs for the current chip manufacturing processes.
- AI could make the pretraining process for the next model more efficient, getting that model sooner.

- AI could improve the pretraining process for the next model to make the model more capable.
- AI could develop ways to fine-tune its own weights to make it more capable.
- AI could modify its harness to make it more capable.

Something these all have in common is that they don't happen instantly. Rather than a continuous process, such as the differential equations behind an object cooling or a pendulum swinging, we instead have something whose capability is increasing in discrete steps.

Such discrete steps are quite common in feedback processes. Even the classic example of a microphone near a speaker gives discrete jumps in volume. When the microphone is switched on it takes a moment for its signal to travel down the wire to the speaker, which then jumps up in volume. This louder sound then has to travel through the air at the speed of sound, before the microphone can register this increase and begin the next cycle. If the microphone were 30 metres from the speaker, the volume would rise in a staircase pattern with each step lasting about a tenth of a second. Because these steps are so brief, we often don't hear them, and continuous models of the feedback are adequate for most purposes.

But the steps in many RSI feedback loops are much longer — potentially months or years. And even more importantly, we've suggested that the RSI feedback loops might be able to produce singular growth. Is that even possible in discrete time or just an artefact of the continuous nature of differential equations?

Taking the discrete nature seriously will change how we see the idealised dynamics of these systems and will change the conditions under which singularities are possible. By the time we're finished, we'll see the possibility of super-exponential, yet sub-singular growth go from being a narrow possibility to a mainline scenario.

We can start by considering the discrete version of the differential equation: the *difference equation*. Let $A_n$ represent the system's ability after $n$ times around the feedback loop. The discrete version of $\dot{A}$ is $\Delta A$, which is defined as:

(14) $\quad \Delta A_n = A_{n+1} - A_n$

In other words, $\Delta A$ is the difference between successive terms in a sequence. Our general autonomous differential equation, $\dot{A} = f(A)$, becomes:

(15) $\quad \Delta A_n = f(A_n)$

Or equivalently (via the definition of $\Delta A$):

(16) $\quad A_{n+1} = A_n + f(A_n)$

In difference equations it is impossible to produce a singularity no matter how quickly $f$ grows. This becomes obvious when one looks at what would be required. How could you have $A_n$ grow without bound prior to some particular $n$? That's only possible if there are infinitely many steps prior to that point or if there is a step where $A_n$ takes an infinite value. But there are only finitely many steps prior to any particular $n$, and all values are finite (they start from a finite level ($A_0$) and each one only adds finitely much to the total, since $f$ is a function from $\mathbb{R}$ to $\mathbb{R}$).

For example, if you try $f(A) = A^2$ (which would give hyperbolic growth using a differential equation) you instead get doubly exponential growth. Or if $f(A) = e^A$, then $A_n$ grows as a tower of exponentials with $n$ levels (tetration). These are very fast rates of growth (clearly super-exponential), yet they have no singularities. For differential equations, rates of growth like these only lived in a very slender zone, but they are entirely generic for difference equations — occurring whenever $f$ is super-linear.

What do these facts about difference equations mean when it comes to RSI? They don't have any direct meaning until we say how the number of feedback cycles ($n$) maps onto time ($t$). Let's create a new model of RSI which is neither a simple differential equation nor difference equation. Instead it will be a difference equation that is embedded into continuous time, which we shall call a *time-embedded difference equation*.

Let $t_n$ be the time at which the *n*th feedback loop is completed and let $T_n$ be the time it takes to go around the feedback loop for the *n*th time. This second parameter ($T_n$) will turn out to be a key quantity when analysing the dynamics of intelligence explosions. I shall call it the *generation time*, via analogy to the parameter of that name in the feedback loops of population growth, disease spread, and nuclear chain reactions. In those subjects, every person/infection/fission produces a random whole number of new persons/infections/fissions, and the generation time is the average amount of time this takes.[7] That corresponds to the amount of time it takes to go around their feedback loops, so I'll use the term 'generation time' even though for some RSI feedback loops there won't be clear discrete generations of AIs.

$T_n$ and $t_n$ are related by the equations:[8]

(17) $\quad T_n = t_n - t_{n-1}$

and

[7] For example, the mean age of the parent at the birth of each of their children.

[8] One could note that the definition of $T_n$ is itself a difference equation: $\Delta t_n = t_{n+1} - t_n = T_{n+1}$. One could therefore think of time-embedded difference equations as a linked system of difference equations for $A$ and for $t$, linked by their index $n$.

(18) $t_n = \sum_{i=1}^{n} T_i$

Let's also define $n_t$ to be the number of cycles around the feedback loop that have been completed by time $t$ (the highest $n$ such that $t_n \leq t$). We can then represent a time-embedded difference equation by the pair of sequences $A_n$ and $t_n$. Together, these enable us to see how $A$ changes over *time*, via:

(19) $A(t) = A_{n_t}$

If every cycle of the feedback loop takes the same time, we'd have $T_n = k$ and $t_n = kn$. If so, the previous remarks about difference equations would apply directly — there are only finitely many loops between any two times, each of which makes only a finite change to $A$, making singularities impossible.

The same is true whenever there is a bound on how short the loop can get. If it never gets shorter than $T_{\min}$ then the growth of $A(t)$ is bounded above by that of a feedback loop with a constant generation time $T_{\min}$.

But what if the generation time decreases towards zero? If it decreases slowly, such as via $T_n = \frac{1}{n}$, then the total time required to perform $n$ loops increases without bound. This implies there are only a finite number of loops between any two times, making singularities impossible. But what if the generation time decreases more quickly? Let's call the time taken to perform infinitely many steps $t_\infty$:

(20) $t_\infty = \sum_{i=1}^{\infty} T_i$

This can be finite. For example, if the generation time shrinks exponentially as $T_n = \frac{1}{2^n}$, then $t_\infty = 1$. In this case an infinite number of loops have been performed within 1 unit of time. As long as $A_n$ increases without bound, this would be a singularity at $t = 1$. So singularities are possible in this model of RSI via a difference equation embedded in time.

What is required to produce them?

First, let's look at the generation time. It has to go to zero fast enough for the infinite sum in (20) to converge. $T_n = \frac{1}{n}$ is too slow, while $T_n = \frac{1}{n^{1+\epsilon}}$ is fast enough. The cut-off for how quickly the denominator of this fraction must grow turns out to be exactly the same as for the blow-up condition we examined earlier. For example, $T_n = \frac{1}{n \log(n) \log(\log(n))}$ is slightly too slow to allow for a singularity, while raising the final factor to any power greater than 1 is sufficient to force the sum to converge, giving a finite $t_\infty$ and thus the opportunity for a singularity. Let's call this condition on how quickly $T_n$ must converge to allow a singularity the *Zeno condition* on $T_n$, as it is precisely the condition for when infinitely many steps can be performed in a finite time.

What about the rate of growth of $A_n$? We've already seen that no growth rate can give a singularity if we only go around the feedback loop finitely many times, but if we get to go around it infinitely many times, then even adding one each time ($f(A_n) = 1$) is enough to produce singular growth. Indeed, even adding smaller and smaller amounts each time around the loop can work, so long as $A_n$ grows without bound. Let's call this the *boundlessness condition* on $A_n$.

For example, if we added just $\frac{1}{n}$ after the *n*th feedback loop, that would be enough to produce unbounded growth and (if the Zeno condition is also met) a singularity. The same is true if we added only $\frac{1}{n\log(n)}$ or $\frac{1}{n\log(n)\log(\log(n))}$. Considered as a function of $n$, the amount that needs to be added each time ($\Delta A_n$) is thus set by exactly the same threshold as $T_n$ (though $\Delta A_n$ mustn't shrink more quickly than this threshold, while $T_n$ must). We can summarise this with the following theorem.

**Singularity Theorem:**

When the growth of $A$ is governed by time-embedded difference equations:

$A(t)$ has a singularity *iff* $T_n$ meets the Zeno condition &
$A_n$ meets the boundlessness condition.

Or equivalently:

$A(t)$ has a singularity *iff* $\sum T_n$ converges & $\sum \Delta A_n$ diverges to $+\infty$

*Proof:* Both conditions are necessary because it doesn't help to grow boundlessly if you can only get through finitely many steps by any particular time, and it doesn't help to have infinitely many steps before some time if the function has a finite bound. But if both conditions are met, then infinitely many steps will happen by $t_\infty$ and these will take $A_n$ beyond every finite level prior to that time.

This theorem cleanly cuts the requirements into two independent thresholds. Success on each is binary: no deficiency on either can be made up by over-performance on the other. We can get a feel for how sharp this is through a pair of examples.

First, consider a feedback loop whose generation time decreases towards zero as $\frac{1}{n}$ and each time it goes through the loop, $A$ gets multiplied by $2^A$. $A_n$ is growing extraordinarily quickly (tetrationally), but because the generation time doesn't meet the Zeno condition, there is no singularity. Second, consider a feedback loop whose generation time decreases towards zero slightly more quickly (as $\frac{1}{n^{1.01}}$), but where only a small and diminishing amount is added to $A$ each time ($+\frac{1}{n}$). This meets both conditions, so blows up to a singularity. The tiny difference in how generation times decreased with $n$ outweighed the radical difference in what happened in each loop.

It is very interesting that in this discrete setting singular growth requires generation time to go to zero, and the rate at which it needs to approach zero is analogous to the rate at which $f(A)$ had to grow in the continuous setting. This suggests that the decreasing generation time is driving the whole effect. One way to see this is that singularities require the gradient of the curve to go to infinity in finite time, and there are two ways to increase gradient: increasing the rise or decreasing the run.

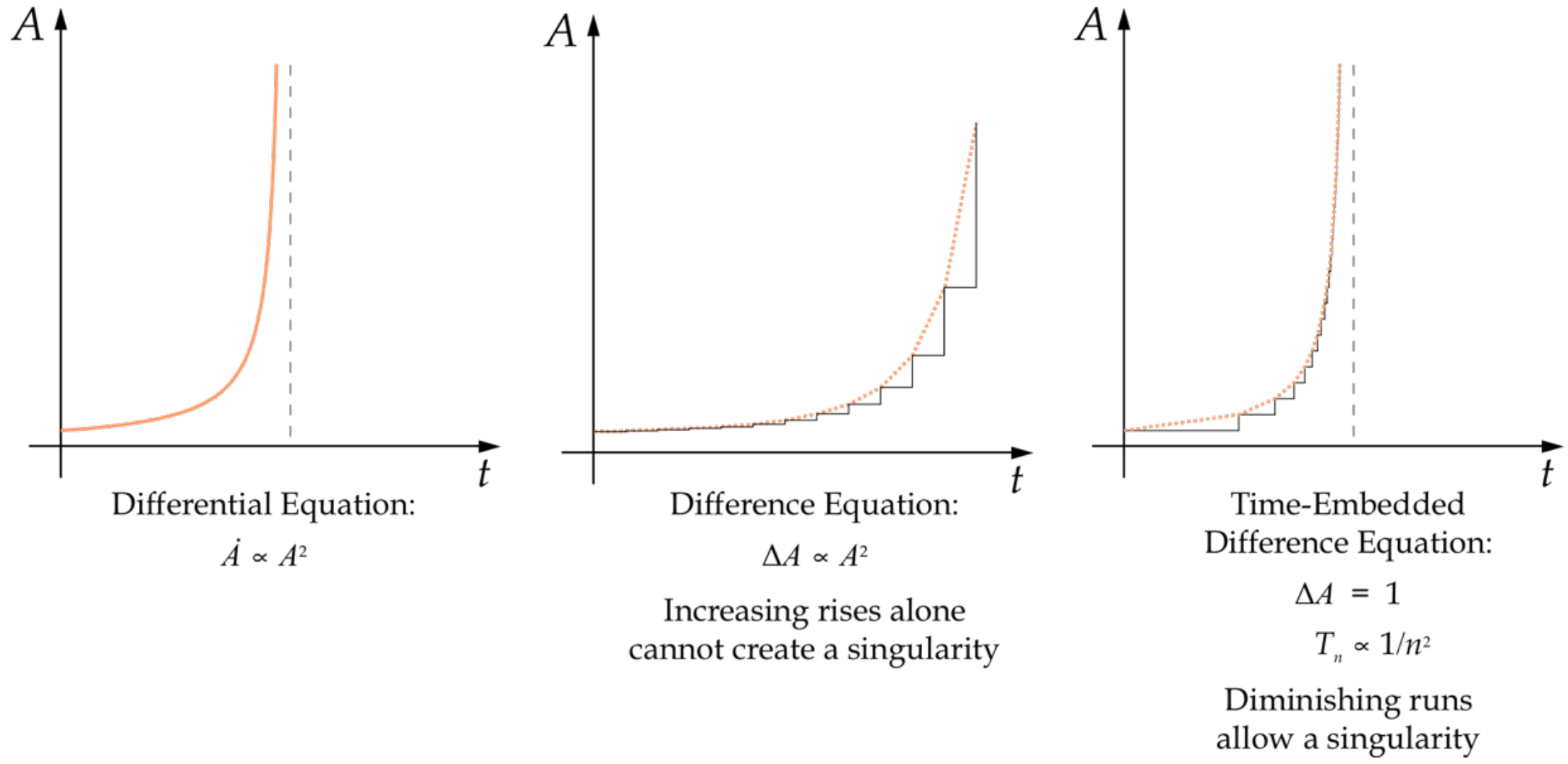

*Figure 1.* Once we take the discrete nature of the feedback loop into account, we see that we can only approach an infinite gradient in finite time by reducing the run in each step towards zero (right).

In continuous models, there is no way to distinguish these. In raw difference equations, one cannot reduce the run and no amount of growth in the rise can produce a singularity. But when we allow for variable-length discrete time steps through a time-embedded difference equation, we see that it is almost entirely about reducing the run. The rise per step doesn't have to increase at all — its only constraint is that it can't *decrease* too quickly.

Another lesson is that a singularity requires the growth of $A$ to be infinitely compounded. We are familiar with the way compounding interest daily leads to a higher growth rate than compounding it annually. Compounding at ever finer intervals converges towards the growth rate set by continuous compounding. But for super-exponential growth, the value of compounding is much greater. If the benefits can only compound finitely many times by a finite point in time, there is no way to achieve a singularity. It is a necessary condition for a singularity that the growth in $A$ is compounded infinitely many times in a finite time interval — which can happen either via a convergent sequence of discrete generation times or a continuous system.

Solomonoff (1985) provides a nice toy model for an intelligence explosion via shortening the generation time. Computing speeds have been improving exponentially over time (a version of Moore's Law). If we reached a point where AI could provide the labour needed to keep Moore's Law running, then every time

Moore's Law doubles computing speeds, it doubles the speed of these digital researchers, halving the time it takes for the next doubling of speeds. So if the first doubling took 2 years, the second would take 1 year, the third half a year… and by the end of 4 years, there would be a singularity — with infinite AI labour and infinite computational efficiency.

Moravec (1999, 2003) developed an improved version of this model, showing that it doesn't even require Moore's Law to be exponential. He first noticed that a quadratic speed-up would suffice, while a linear speed-up would not. Then he found his way to the peculiar threshold we've seen so many times: so long as the *n*th generation machines are faster than the first generation by a factor greater than $n \log(n) \log(\log(n)) \ldots$ this approach produces a singularity.

Of course, there are still great obstacles making this impractical. Moore's Law has required increasing amounts of labour to keep it going (due to fishing-out and stepping-on-toes effects), as well as increasing amounts of capital. Moreover, there will be ultimate physical limits to how far it can go, and there are practical limits to how quickly a new generation of faster chips can be produced (especially when a novel method is required). But it is still an instructive model for an efficiency-driven intelligence explosion. While Solomonoff and Moravec modelled it with differential equations, the presence of an explicit generation time that tends towards zero allows this model to work just as well in time-embedded difference equations.

Eth and Davidson (2025) considered something like this when they explored AI systems that improve their own training time. Suppose an AI system could make sufficient tweaks to its architecture and training process that it slightly improved the speed of training and inference for the next generation of AI systems. If this could be repeated indefinitely, then the generation time (to design and train the next generation) would keep falling. So long as the speed improvements are above the familiar threshold ($n \log(n) \log(\log(n)) \ldots$), one would get singular growth, leading to unlimited AI labour by some finite time. Of course, Eth and Davidson don't suggest that we can actually make an indefinite series of such improvements. Instead, one would expect the generation time for training the next generation of models to bottom out at some unyielding finite limit, prematurely ending the period of singular growth.

In general, it seems highly unlikely that generation times can be brought arbitrarily close to zero. This provides an important kind of barrier to singular growth.

It is essential in all these arguments that we are measuring generation time itself, and not the related concept of *doubling time*:

(21) $$D_n = \frac{T_n}{\log_2\left(\frac{A_n}{A_{n-1}}\right)}$$

Doubling time is often useful as it takes into account how much a feedback loop contributes as well as how long it takes. For instance, a loop that takes one second and doubles its input has the same doubling time as one that takes two seconds and quadruples its input. A key advantage of doubling time is that it is also defined for continuous processes, while generation time is not.

However, this combining of the loop's duration and impact turns out to be doubling time's downfall. If $A_n$ grows very quickly (e.g. doubly exponentially) while the generation time stays constant, then the doubling time will rapidly decline to zero. But we've seen that since generation time is constant, there can't be singular growth no matter how quickly $A_n$ grows or $D_n$ shrinks. Doubling time shrinking towards zero is a useful threshold for defining super-exponential growth, but it is only generation time that can set the threshold for singular growth.

It is worth noting that one could also model the length of feedback loops via *delayed differential equations* rather than difference equations. In the equation $\dot{A} = f(A)$ we'd been suppressing the dependence on time. We could have equally well written $\dot{A}(t) = f(A(t))$. We can change this to a delayed differential equation where the current value of $\dot{A}$ depends on the value of $A$ at a time $T$ units earlier (to account for the generation time): $\dot{A}(t) = f(A(t - T))$. Like the difference equation, this delayed differential equation with a fixed delay can produce super-exponential growth but can't produce a singularity.

And we can make an analogue to the time-embedded difference equation, by allowing the feedback loop length to change with time: $\dot{A}(t) = f(A(t - T(t)))$. Like the time-embedded difference equation, this general delayed differential equation *can* produce a singularity, but only when $T$ shrinks towards zero sufficiently quickly as $t \to t^*$. This gives a continuous model with similar dynamics. I find it slightly less accurate (losing track of the discrete nature of the feedback loop) and slightly harder to work with, but other researchers may find it useful.

We are now able to zoom out and take another look at when growth from feedback loops is singular versus merely super-exponential:

- For differential equations of the common restricted form, $\dot{A} = kA^r$, all super-exponential growth is singular growth.
- For differential equations of the more general form, $\dot{A} = f(A)$, we see that there is also a narrow range of super-exponential growth without a singularity.
- When we take the discrete nature of feedback loops into account with difference equations of the form $\Delta A_n = f(A_n)$, all super-exponential growth is sub-singular.
- When we allow for changing generation times using time-embedded difference equations, we can again get super-exponential growth in singular and sub-singular varieties. But compared with the setting of differential equations, singular growth is now much harder to achieve due to the requirement that the length of each feedback loop shrinks towards zero. Super-exponential growth without a singularity is no longer a curiosity — it is the default.

**Intelligence Measures**

When using a differential or difference equation to model RSI, it is remarkably unclear what $A$ should represent. Should it represent intelligence, or should we sidestep that and measure the computational efficiency of that intelligence? If we choose to represent intelligence itself (and thus try to model an actual *intelligence* explosion), we run into two further problems.

First, it is widely recognised that there is no generally agreed conception of intelligence to measure. Even among those who agree that intelligence is a real and important thing, there is no consensus on which thing it is.

AI research tries to sidestep this by measuring a wide variety of different capabilities via benchmarks. These measure what fraction of a set of tasks related to that capability the system can successfully solve. There is a common feeling that achieving artificial general intelligence (AGI) will require roughly human performance on most such capabilities, but there is some dissent. For example Chollet (2019) argues that intelligence is really the ability to efficiently learn a wide variety of capabilities. On his view, intelligence isn't measured in terms of capability, but *capability per unit input* (where inputs could include training data, information embedded in the priors, training compute, inference compute etc.). Even if there were agreement on whether intelligence is a measure of capability or capability per unit input, one would need further agreement on how to weight the different kinds of capabilities into an overall linear ordering for $A$.

There is also a second problem, which is much less widely recognised. Even if we could agree on the kind of thing being measured (e.g. we could agree on a linear ordering of all AI systems according to intelligence) it is very unclear what cardinal structure this should have — how to assign numbers to those systems.

For example, when measuring the capability of AI systems at playing a game such as chess or Go, researchers often use Elo scores. But Elo scores are really just the log of a more fundamental measure from the Bradley-Terry model which assigns each player a strength such that the odds ratio of a player beating another is simply the ratio of their strengths. Elo is just the log of this strength (with some arbitrary constants to help its numbers match an earlier chess ranking system).

This causes confusion in the literature when papers claim that in contrast to some other domains, chess AI is only improving linearly over time. Chess improvements have been roughly linear when measured in Elo, but the headline claim could equally be that chess-playing AI is improving exponentially over time (when measured in the improvement in the odds-ratio of beating a player of fixed strength). In this case, it is unclear whether progress in chess AI is best described as exponential or linear.

Overall measures for AI progress suffer from the same issue. They could be exponential on one fairly natural scale and linear on another. Or they could be convex (with increasing returns) on one scale while concave (with diminishing returns) on another.[9] Given there is often very little discussion or agreement on which measure is the more natural one (or whether there is even a fact of the matter about that) there is a big problem of measure-dependence for attempts to track progress in AI.

How do these issues affect the dynamics of RSI — or our ability to measure and track those dynamics?

Let's start with a simple example where $A$ is a measure of intelligence and $B$ is another measure where $B = \log(A)$. For the differential equation model, where $\dot{A} = f(A)$, we will also have $\dot{B} = g(B)$, where $g$ is a slower growing function than $f$. For example, when $\dot{A} = kA$, $\dot{B} = k$. These are consistent (when the measure $A(t)$ grows exponentially, it makes sense that the measure $B(t)$ is growing linearly) but they show that the same rate of progress of the observable phenomenon in the world can correspond to different functions $f$ in the differential equation.

However, there is an important invariant: $f(A)$ grows fast enough to produce a singularity if and only if $g(B)$ grows fast enough to produce a singularity. i.e. $f$ will meet the blow-up condition (of growing faster than $A \log(A) \log(\log(A))$ etc.) if and only if $g$ also does. It is easy to see this must be true because when some quantity grows without bound, the log of that quantity also grows without bound. So if one grows without bound by time $t^*$ the other must too.

Let's define two measures as *similar* when either measure growing without bound implies the other does too. When $A$ and $B$ are similar, then either both $f$ and $g$ meet the blow-up condition or neither do. Therefore, if we are trying to detect singular growth of an RSI system in terms of how $\dot{A}$ is increasing with $A$, the threshold is at the same place (roughly $A \log(A) \log(\log(A))\ldots$) for a wide range of different ways of measuring that system's capabilities.[10]

Something very similar is true if we model RSI via time-embedded difference equations. In this case, the only necessary condition on $A_n$ to achieve a singularity is that it grows without bound. And if $A_n$ and $B_n$ are similar, then they either both satisfy this condition or neither does. So the threshold for how much they need to improve on the $n$th feedback loop ($n \log(n) \log(\log(n))\ldots$) is the same for both, and we don't need to worry about which of these capability measures we are using when testing for a singularity.

[9] For example, it may be that on a linear measure their score is rising as $t^{10}$ while on a logarithmic measure it is only rising as $10 \log(t)$.

[10] The same is *not* true for the threshold of super-exponential growth, since $A(t)$ could grow super-exponentially while its log, $B(t)$, doesn't.

But some measures are not similar to each other. For example, when the full range of measure $A$ only maps into a finite interval in measure $B$, then $A$ going to infinity doesn't entail $B$ going to infinity and you could get singular growth in $A$ without singular growth in $B$. This could easily happen if $A$ represented open-ended progress in some domain of intelligence while $B$ was a broader measure that included more domains. In such a case, there would be no contradiction in $A$ having singular growth without $B$ having it too. The time $t^*$ would simply be the time at which problems of the domain $A$ was measuring are effectively solved.

This might be the case for measures of $A$ like computational efficiency or 'effective compute'. Many problems would be solved if we let an AI system have unlimited effective compute (for training and/or inference), but it isn't clear such a system would excel at all kinds of intellectual tasks. For example, even if you had unlimited compute, it isn't clear that current architectures and training environments would allow a system to succeed in domains which are lacking a clean algorithmic way of rating the quality of the outputs. Much of the recent progress in AI has relied on reinforcement learning with verifiable rewards (RLVR), but when there are no verifiable rewards, performance may plateau. If so, even a completed singularity in measures like efficiency and effective compute needn't imply that the system is more capable than a human across the board after time $t^*$.

In physics, scientists distinguish between a *coordinate singularity* and an *essential singularity*. The former exists when the singularity is an artefact of how things are being measured. For example, when Schwarzschild (1916) published his mathematical description of a black hole there was a mathematical singularity at the black hole's event horizon. It took decades of scientific work before physicists were able to establish that this was a mere artefact of the coordinates being used (whereas the singularity at the centre of the black hole was not fixable by a change in coordinates).

We can adopt this distinction when studying RSI. For example, suppose we are measuring $A$ by the *mean time between failures (MTBF)*: how long the system can go before making a mistake while completing tasks of a certain kind. This is a standard measure in engineering and a relevant measure of capability for many real-world uses. In this case, one might see $A$ rising towards a singularity. But consider an alternative measure $B$ which is the percentage of times the system gets the right answer. If $A$ goes to infinity at time $t^*$, that just means $B$ reaches 100% at $t^*$. While a lot of systems never quite reach 100% reliability, there is nothing impossible or paradoxical about doing so. Thus in this example, $t^*$ is a coordinate singularity in measure $A$, but not in measure $B$, and is not an essential singularity.

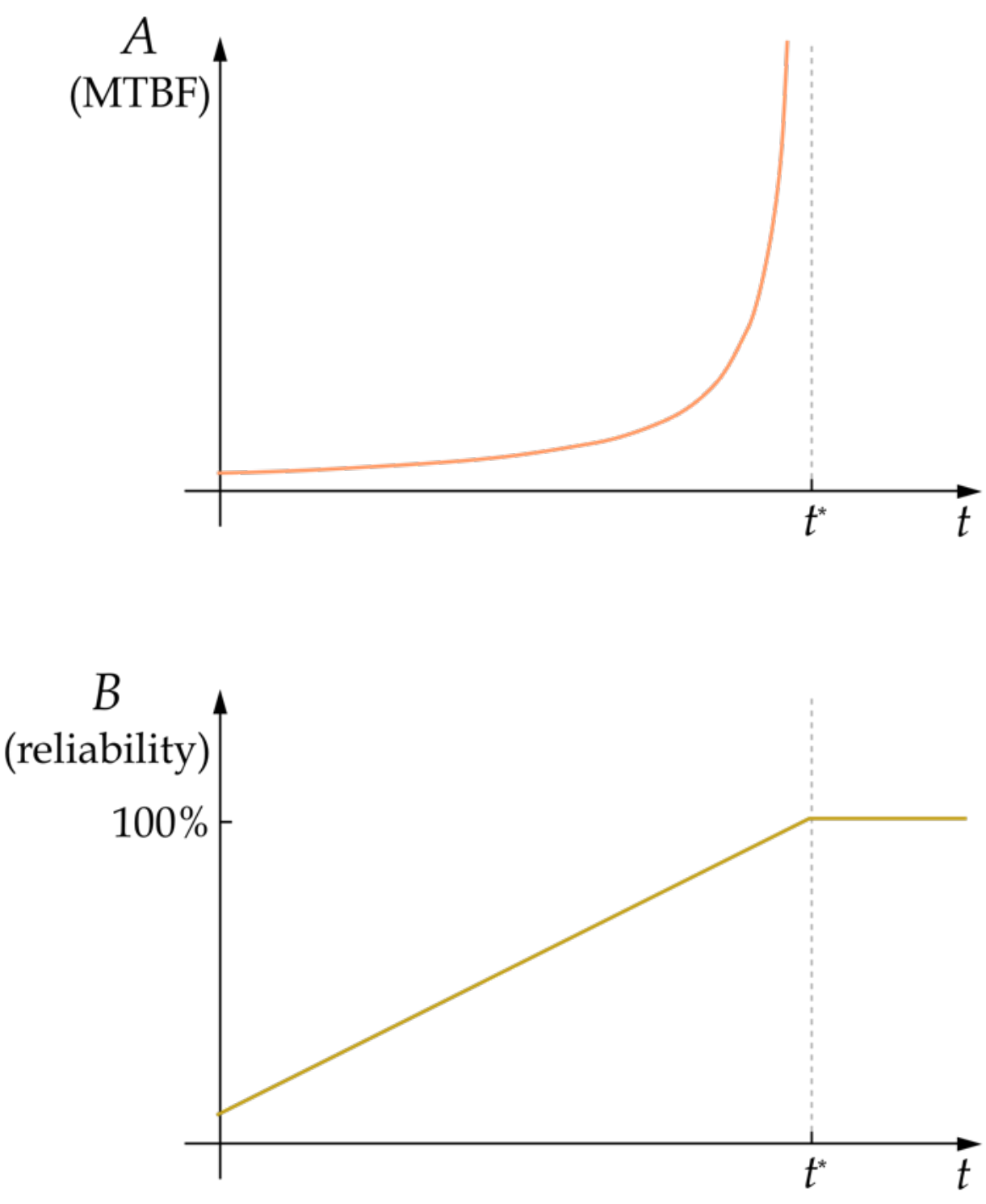

*Figure 2.* How two different measures of the same AI capabilities evolve over time. Measure $A$ (mean time between failures) has a singularity at time $t^*$, while measure $B$ (reliability) simply reaches 100% at that point.

Note that a very widely used measure of general AI capability — the METR time horizon (Kwa et al. 2025) — is somewhat like a mean time between failures. Recent years have seen exponential (or perhaps super-exponential) growth in time horizons for AI systems completing the kinds of routine professional computer-use tasks in the benchmark. But even if the time horizon went to infinity[11], it isn't clear the system would be as intelligent as a human. It may still lack various cognitive skills (such as continual learning, sample efficiency, or creativity), which are important but not being tested. And there may still be substantial room for other AI systems to be much more capable, even at the skills that were being tested — systems that could solve much more challenging computer-use problems using much less compute. An infinite METR time horizon would just mean that the AI system can do the kinds of task in the benchmark with perfect reliability (and unbounded duration).

[11] There are practical reasons why the METR horizon length may not be able to go to infinity— the tasks included in the benchmark only go up to 16 hours of human time and METR explicitly say their current version can't be used to estimate the horizon lengths beyond that. But I'm pointing to a theoretical issue — even if they had endless capacity to create longer tasks and endless immortal human baseliners, to enable arbitrarily high measurements, this is the kind of measure where time horizons diverging to infinity correspond to accuracies converging to 100% on a fairly narrow range of cognitive tasks, rather than infinite intelligence.

So even though most people studying intelligence explosions assume that the capability measure $A$ won't actually go to infinity in finite time, it wouldn't be absurd or unphysical for it to do so. It depends on what measure it is. Some measures could go to infinity without incident. If such measures could also satisfy the key differential equation (or time-embedded difference equation) then one could get a completed intelligence explosion — $A$ would have gone to infinity in finite time.

I'm not quite sure what we should think about this. It really depends on the nature of the measure, and this result is much easier to achieve on narrow measures — those that can rise unboundedly even when many key cognitive abilities are lacking. But measures like mean time between failures or METR time horizons seem unlikely to lead to rapidly decreasing generation times — what goes to infinity is a measure of duration, not a measure of speed. By showing that a completed singularity is theoretically possible, I'm mainly trying to caution people about the behaviour of certain intelligence measures, rather than trying to suggest we will genuinely have unbounded progress in intelligence in a finite time.

Let's now turn to look at what happens when an intelligence explosion can't get to infinity.

**Going Finite**

We've now seen how super-exponential growth can be cleanly divided into singular and sub-singular varieties, with the singular kind appearing very difficult to achieve (once we take the discrete nature of feedback loops into account). And we've seen that the rates at which $f(A)$ has to grow or diminish are independent of which measure is used (so long as the relevant measures are 'similar' to each other).

But all of these claims rely on a clean mathematical model where growth is classified by its asymptotic behaviour as $t$ or $A$ approaches infinity. In the real world, this model may very well break at some finite value of $t$ or $A$, giving increasingly inaccurate results thereafter. That's a big problem for this kind of asymptotic analysis — as it relies on the infinite domain (or range) to produce its clean classifications. I hope that these divisions which are natural in the idealised setting will be natural in the real-world application. This is often the case[12], though it isn't guaranteed.

[12] For example, the class of functions computable by Turing machines is of immense use in computer science, even though it collapses to be the same as the functions computed by finite state machines (or look-up tables) if we restrict ourselves to finite settings. Similarly, the asymptotic analysis of time complexities of algorithms is very useful even though it disappears if we were to restrict ourselves to algorithms whose inputs are bounded by the size of the observable universe.

There are a variety of finite limits we might expect to run into, which could force the clean mathematical model to break at different points, and in different ways.

The most widely discussed is a *ceiling* on intelligence: a horizontal line at some level $A^*$, which $A(t)$ cannot cross. In the study of feedback systems, people often say that $A$ *saturates* as it approaches such a horizontal asymptote. It is generally thought that this will happen for RSI.

There are many different kinds of ceiling at which the AI's intelligence (or efficiency) may saturate:

- *Limits of intelligence itself* — Even an optimal reasoner would be neither omniscient nor omnipotent. It would need to perform experiments to gain knowledge, could still be beaten at unbalanced games, and may face intractable prediction problems in chaotic or agentic domains.
- *Limits of intelligence per unit resource* — Even if we have the optimal algorithms and hardware, our solar system has only one star out of the 200,000,000,000 in our galaxy, and growth beyond our system is slow and cubic. If a galactic superintelligence would be more intelligent than a stellar superintelligence, it will be a long time before we could reach that higher level.
- *Limits of the hardware paradigm* — Even an optimal silicon chip may be far below the physical limits of compute per unit resource.
- *Limits of the algorithmic paradigm* — Even an optimal neural network may be far below the best intelligence that could be achieved with that amount of compute.
- *Limits of training data* — The training data we have (and could acquire during RSI) is lacking a lot of information on many domains (especially non-verbalisable information and contextual information). So we could have a very intelligent system that is limited by inability to train certain skills.
- *Earlier limits* — The ascent may stall out before any of the above due to limitations inherent in the starting AI.

Discussions of RSI sometimes tacitly assume that an intelligence explosion would only stop when it reached the limits of intelligence itself, and then use assumptions like omniscience or perfect rationality to model its behaviour. But an explosion might saturate significantly below this level, with important consequences for predicting post-explosion capabilities and for understanding whether all intelligence explosions need end at the same intelligence level.

The manner in which super-exponential growth slows down as it approaches its limit might be analogous to the way exponential growth runs out of steam. While the simple equation for exponential growth is $\dot{A} = kA$, in reality there is usually an additional dampening term that starts small but grows to dominate for large $A$. For example if we subtract a quadratic term that has been shrunk down by a large factor ($K$), this gives the differential equation for logistic growth, $\dot{A} = kA - \frac{kA^2}{K}$, where $A(t)$ approaches a horizontal asymptote at height $K$. We could use the same model for super-exponential growth: $\dot{A} = f(A) - \frac{f(A)^2}{K}$. Or more generally, we could assume $\dot{A} = f(A) - g(A)$, where $g(A_0) \approx 0$ and $g(A) < f(A)$ until some crossover point, $A^*$. This will produce a horizontal asymptote at height $A^*$.[13]

This dampening force could come from running out of room for improvement as a system approaches some form of perfection (such as *optimal intelligence per unit resource*), or it could come from something like straining under the growing size or complexity of the system. The latter could stop the ascent before reaching any kind of optimal system.[14]

As well as a ceiling on $A$, one could also run into upper limits on how quickly it can increase — either in absolute terms (such as a limit on $\dot{A}$) or percentage terms (such as a limit on $\frac{\dot{A}}{A}$). Rather than making $A(t)$ converge to a horizontal asymptote, a maximum gradient would make $A(t)$ converge towards a diagonal line, while a maximum growth rate would make $A(t)$ converge towards an exponential.

What could produce limits of these kinds?

One mechanism I find particularly likely is if there is a floor on the achievable generation time, $T_n$. It seems very unlikely that generation times can be brought arbitrarily close to zero. There are many kinds of feedback loop that could contribute to RSI, ranging from decades (e.g. designing a successor to EUV lithography) to months (e.g. designing better pretraining) to seconds (e.g. designing better scaffolds). While some are extremely short, those only capture a tiny fraction of the pipeline of what makes for better AI R&D, so it seems likely they would quickly saturate if the other aspects of AI improvement were unchanged. For example, putting a fixed agent in a scaffold that has the agent repeatedly redesign that very scaffold might make some improvements, but without changing the model itself, it seems very unlikely to take off towards infinity. And while continued RSI might be able to reduce any of these generation times by a substantial factor, each one seems likely to run into constraints where a certain minimal amount of time has to pass in order to make an improvement.

---

[13] Here I've been taking the approach of treating $f(A)$ as a clean mathematical function that only captures part of the dynamics, so requires a correction term. A different way to view things is to let $f(A)$ represent the actual messy real-world connection between $\dot{A}$ and $A$ (i.e. the empirical function you would graph as you measure an intelligence explosion). On this approach you don't need to subtract a second function, but instead would talk about whether the real $f(A)$ eventually falls to zero at some level of $A$, creating a horizontal asymptote at that height.

[14] Standard models based on semi-endogenous growth theory model diminishing returns in a scale-free way, such that diminishing returns alone can't come to halt the ascent of $A$. But this is just a byproduct of their simple power-law model. It is very easy to get horizontal asymptotes in more flexible differential equations.

What would a lower bound on the achievable generation time do? If $T_n$ approached some minimal value ($T^*$) while $A_n$ grew by a fixed amount each loop ($\Delta A_n = k$), then $A(t)$ would approach a linear rate of increase. If $A_n$ instead grew by a fixed proportion each loop ($\Delta A_n = kA_n$), then $A(t)$ would approach an exponential rate of increase.

I think the second of these looks quite plausible. If so, we might see multiple phases of an explosion:

0. The initial exponential phase when the doubling time of $A$ is driven by human-only research.
1. Increasing amounts of RSI drive the generation time down towards machine speeds, so the growth rate of $A$ climbs from its human-only rate towards some very high fully automated rate. Because this phase involves a continuously increasing growth rate, it is super-exponential in shape.[15]
2. Fully automated RSI continues for a while at this faster exponential, but as $A$ climbs, it starts to saturate. This phase would be exponential in shape.
3. As it approaches its inflection point and subsequent horizontal plateau, the trajectory has substantially departed from its exponential form and is revealed to be a logistic.

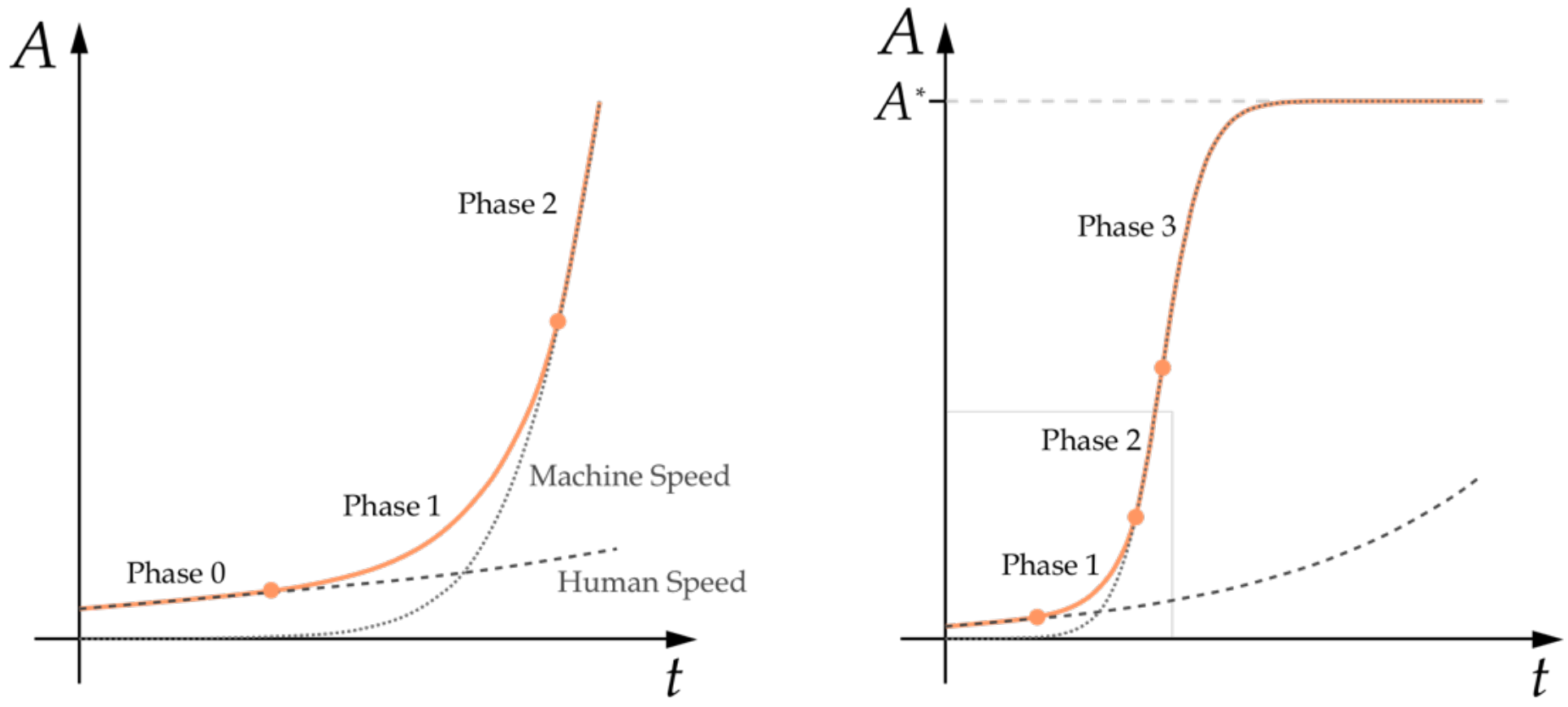

*Figure 3.* Phases of an intelligence explosion. The left diagram shows the first phases, with $A(t)$ moving from a human-speed exponential through a period of super-exponential

[15] Though note that the amount by which the growth rates are rising per year slows down towards the end, so any particular clean super-exponential shape would only fit this phase until this point of inflection of growth rates.

growth to converge on a machine-speed exponential. The right diagram zooms out to show the final phase, where $A$ saturates at some high level $A^*$.

This trajectory involves two kinds of finite limits on the idealised process — first on generation time (and thus growth rate of $A$), then on the absolute level of $A$. It shows how we could have a super-exponential process (which would be worth modelling as such), yet it runs out of steam in two distinct ways. On this view, the mathematical models of super-exponential growth could be a good fit (and thus predictive) for the period where generation time is reducing towards zero in some clean way (e.g. 10% shorter each loop), but then cease to be a good model as we enter phase 2.

I've been assuming that $A$ would be improving by a constant factor each time round the loop, guaranteeing super-exponential and logistic phases, but versions with other functions for $f(A_n)$ are also possible and can still show distinct phases if generation time saturates prior to intelligence level saturating.

**Conclusions**

My aim in this paper has been to improve our theoretical understanding of intelligence explosions — looking at the most explosive possibilities (super-exponential and singular growth) and asking what conditions are required to produce them.

We first looked at models based on differential equations and saw that the key threshold for super-exponential growth of some measure $A$ is when $\dot{A}$ grows superlinearly in $A$. But the threshold for singular growth is somewhat higher — $\dot{A}$ has to grow slightly faster than all functions of the form $A\log(A)\ \log(\log(A))\ldots$ This allows for the possibility of ending up with growth that is super-exponential yet doesn't approach a finite time singularity.

We then saw that when you take the duration of the feedback loop into account, singular growth becomes much harder — requiring this generation time to approach zero (and quickly enough). Indeed, this model helps us see that singular growth is more about reducing the generation time than it is about increasing how much each loop contributes. And the dynamic that produces singular growth is simply the combination of the Zeno condition and the boundlessness condition. On this more realistic model, growth that is super-exponential without being singular looks much more likely. It is what you get whenever you can make $\Delta A$ grow super-linearly in $A$, yet can't reduce the generation time all the way towards zero. So analysis of explosive growth needs to be careful to distinguish singular growth from merely super-exponential growth.

For example, suppose we saw increasing monthly growth rates in some key intelligence measure: 10% growth, then 20%, then 30%. From the economics-inspired literature, we might have thought that since this is super-exponential, it must be hyperbolic growth. But we now know increasing growth rates are not the signature of singular growth. (If the pattern above continued, $A$ would 'merely' be growing as $e^{0.05t^2}$.)

The importance of generation time to the dynamics of intelligence explosions suggests that generation times need to be carefully measured and tracked. It may be a good policy idea to require frontier labs to report their current generation times — especially those for pre-training and for RLVR post-training.

Two of the most important implications for the measurement of an intelligence explosion are negative. When we move beyond the simple model of $\dot{A} = kA^r$, you cannot tell whether a process will explode or not based on its returns over a finite period. This is because the condition for having a singularity is a global property of the relationship between $\dot{A}$ and $A$ (the blow-up condition), rather than a local property such as elasticity > 1. The same is true when modelling it with time-embedded differential equations or delayed differential equations, where the Zeno and boundlessness conditions are inherently global. In all cases, deficiencies in growth somewhere can be made up for by more extreme growth elsewhere.

So without strong assumptions about the functional form, you can neither rule out nor confirm that an explosion is under way based on local measurements. So while we should probably take measurements of elasticity > 1 for $f(A)$ over some range of $A$ as evidence in favour (and elasticity < 1 as evidence against) that evidence is limited. We should also consider other forms of evidence such as how the elasticity has been changing, or how quickly — and how far — the generation time can be reduced.

While it is not unique to my modelling, I also want to stress that all the kinds of feedback loops we've explored have behaviours that are exquisitely sensitive to fine differences near the borderlines of their behaviours. In contrast, the empirical measurements of intelligence that we can perform are all quite rough and noisy. So if the measured values suggest the growth of $f(A)$ or the shrinking of the generation time is anywhere near this border, it will be extremely hard to empirically determine the future behaviour. These limitations are important for companies setting their internal evaluations and for the prospects of regulation.

After examining what produces singular growth, we also looked at the choice of how we measure intelligence. We saw that whether intelligence is growing super-exponentially often depends on the nature of our measure, while whether it is growing towards a singularity is much less dependent on it. This fact means it isn't crucial to decide whether to use some measure or the log of that measure — you would look for the same signature of how $\dot{A}$ is growing with $A$ in both cases. But when one measure can rise to infinity without the other following suit, the choice of measure really matters. In particular, some popular measures like METR time horizons are probably not the right tool for the job, since they can go to infinity without bringing other measures of intelligence along with them.

And we looked at how various finite limits might bite first, preventing truly singular growth. As well as the familiar idea of intelligence saturating at some maximal level (or maximal given various constraints) a minimal generation time could also derail the models of singular growth far before we reach that point.

Finally, while I've argued that singular growth is harder than we may have thought, that doesn't mean RSI is safe or that AI R&D will move at a manageable pace. RSI might be able to speed AI R&D up to dangerously fast speeds even just with a linear speed-up. For example, if the human-only trajectory were $A(t)$ and RSI sped this up to $A(10t)$, we'd be getting a decade of human-only progress each year, introducing many of the dangers — even without any change in the fundamental shape of the curve.

## Appendix: Table of Rates of Growth

| $f(A)$ | $A(t)$ | Singularity? |
|---|---|---|
| … | … | No |
| $e^{-A}$ | $\sim\log(t)$ | No |
| $A^{-999}$ | $\sim t^{0.001}$ | No |
| $A^{-1}$ | $\sim t^{0.5}$ | No |
| $k$ | $\sim kt$ | No |
| $A^{0.001}$ | $\sim t^{1.001}$ | No |
| $\sqrt{A}$ | $\sim t^{2}$ | No |
| $A^{0.999}$ | $\sim t^{1000}$ | No |
| $A$ | $\sim e^{t}$ | No |
| $A\log(A)$ | $\sim e^{e^{t}}$ | No |
| $A\log(A)\log(\log(A))$ | $\sim e^{e^{e^{t}}}$ | No |
| … | … | … |
| $A\log(A)\log(\log(A))^2$ | $\sim e^{e^{\left(\frac{1}{t^*-t}\right)}}$ | Yes |
| $A\log(A)^2$ | $\sim e^{\left(\frac{1}{t^*-t}\right)}$ | Yes |
| $A^{1.001}$ | $\sim\frac{1}{(t^*-t)^{1000}}$ | Yes |
| $A^2$ | $\sim\frac{1}{t^*-t}$ | Yes |
| $A^3$ | $\sim\frac{1}{\sqrt{t^*-t}}$ | Yes |
| $A^{1001}$ | $\sim\frac{1}{(t^*-t)^{0.001}}$ | Yes |
| $e^{A}$ | $\sim -\log(t^*-t)$ | Yes |
| … | … | Yes |

*Table 1*. The rates of growth of $A$ over time produced by different rates of return $f(A)$ in the differential equation $\dot{A} = f(A)$. Note the unusually 'mild' singular growth from $f(A) = A\ log(A)^2$, which is faster, though less 'explosive' than any hyperbolic growth, and the unusually severe singular growth, from $f(A) = e^A$, which is lags behind any hyperbolic growth, but explosively catches up at the final instant.

## References

Philippe Aghion, Benjamin F. Jones, and Charles I. Jones. (2017). 'Artificial Intelligence and Economic Growth', NBER Working Paper 23928.

Marcin Andrychowicz, Misha Denil, Sergio Gomez, Matthew W. Hoffman, David Pfau, Tom Schaul, Brendan Shillingford, Nando de Freitas. (2016). 'Learning to learn by gradient descent by gradient descent', https://arxiv.org/abs/1606.04474

Anthropic. (2026). System Card: Claude Mythos Preview [System Card]. https://www.anthropic.com/claude-mythos-preview-system-card

Nicholas Bloom, Charles I. Jones, John Van Reenen, and Michael Webb. (2020). 'Are Ideas Getting Harder to Find?', *American Economic Review*, 110(4):1104–44.

Nick Bostrom. (2014). Superintelligence: Paths, Dangers, Strategies, Oxford University Press.

David J. Chalmers. (2010). 'The Singularity: A Philosophical Analysis', *Journal of Consciousness Studies* 17:7-65.

Alan Chan, Ranay Padarath, Joe Kwon, Hilary Greaves, Markus Anderljung. (2026). 'Measuring AI R&D Automation', https://arxiv.org/abs/2603.03992

François Chollet. (2019). 'On the Measure of Intelligence'. https://arxiv.org/abs/1911.01547

Tom Davidson. (2025a). 'How Can AI Labs Incorporate Risks from AI Accelerating AI Progress Into Their Responsible Scaling Policies?'. https://www.forethought.org/research/how-can-ai-labs-incorporate-risks-from-ai-accelerating-ai-progress-into

Tom Davidson. (2025b). 'Will the need to retrain AI models from scratch block a software intelligence explosion?'. https://www.forethought.org/research/will-the-need-to-retrain-ai-models

Tom Davidson, Jean-Stanislas Denain, Pablo Villalobos, and Guillem Bas. (2023). 'AI capabilities can be significantly improved without expensive retraining.' https://arxiv.org/abs/2312.07413

Tom Davidson, Rose Hadshar, and Will MacAskill. (2025). 'Three Types of Intelligence Explosion' https://www.forethought.org/research/three-types-of-intelligence-explosion

Tom Davidson, Basil Halperin, Thomas Houlden, and Anton Korinek. (2026). 'When Does Automating AI Research Produce Explosive Growth? Feedback Loops in Innovation Networks', NBER Working Paper No. 35155.

Tom Davidson and Tom Houlden. (2025). 'How quick and big would a software intelligence explosion be?'. https://www.forethought.org/research/how-quick-and-big-would-a-software-intelligence-explosion-be

Ege Erdil, Tamay Besiroglu, and Anson Ho. (2024). 'Estimating Idea Production: A Methodological Survey', *SSRN*. http://dx.doi.org/10.2139/ssrn.4814445

Daniel Eth and Tom Davidson. (2025). 'Will AI R&D Automation Cause a Software Intelligence Explosion?'. https://www.forethought.org/research/will-ai-r-and-d-automation-cause-a-software-intelligence-explosion

A. Fawzi, M. Balog, A. Huang et al. (2022). 'Discovering faster matrix multiplication algorithms with reinforcement learning'. *Nature* 610:47–53. https://doi.org/10.1038/s41586-022-05172-4

I. J. Good. (1965). 'Speculations Concerning the First Ultraintelligent Machine,' In F. Alt & M. Rubinoff, *Advances in Computers* (volume 6). Academic Press.

Demis Hassabis, Dario Amodei, and Zanny Minton Beddoes. (2026). *The Day After AGI*. World Economic Forum [Panel]. https://www.weforum.org/meetings/world-economic-forum-annual-meeting-2026/sessions/the-day-after-agi/

Charles I. Jones. (1995). 'R&D-Based Models of Economic Growth,' *Journal of Political Economy* 103(4):759–84.

Andrej Karpathy. (2026). 'Autoresearch'. https://github.com/karpathy/autoresearch

Michael Kremer. (1993). 'Population Growth and Technological Change: One Million B.C. to 1990', *The Quarterly Journal of Economics* 108(3):681–716.

Daniel Kokotajlo and Eli Lifland. (2025). 'Takeoff Forecast' https://ai-2027.com/research/takeoff-forecast

Raymond Kurzweil. (2001). 'The law of accelerating returns'. https://www.writingsbyraykurzweil.com/the-law-of-accelerating-returns

Ray Kurzweil, Vernor Vinge, and Hans Moravec. (2003). 'Singularity math trialogue'. https://www.thekurzweillibrary.com/singularity-math-trialogue

Thomas Kwa, Ben West, Joel Becker, Amy Deng, Katharyn Garcia, Max Hasin, Sami Jawhar, Megan Kinniment, Nate Rush, Sydney Von Arx, Ryan Bloom, Thomas Broadley, Haoxing Du, Brian Goodrich, Nikola Jurkovic, Luke Harold Miles, Seraphina Nix, Tao Lin, Neev Parikh, David Rein, Lucas Jun Koba Sato, Hjalmar Wijk, Daniel M. Ziegler, Elizabeth Barnes, Lawrence Chan. (2025). 'Measuring AI Ability to Complete Long Software Tasks'. https://arxiv.org/abs/2503.14499

Eli Lifland, Brendan Halstead, Alex Kastner, and Daniel Kokotajlo. (2026) *AI Futures Model*. https://www.aifuturesmodel.com

Hans Moravec. (1999). ‘Simple equations for Vinge’s technological singularity’. https://frc.ri.cmu.edu/~hpm/project.archive/robot.papers/1999/singularity.html

Hans Moravec. (2003). ‘Simpler equations for Vinge’s technological singularity’. https://frc.ri.cmu.edu/~hpm/project.archive/robot.papers/2003/singularity2.html

Anders Sandberg. (2013). ‘An Overview of Models of Technological Singularity’. In *The Transhumanist Reader* (eds M. More and N. Vita-More). https://doi.org/10.1002/9781118555927.ch36

Karl Schwarzschild. (1916). ‘Über das Gravitationsfeld eines Massenpunktes nach der Einsteinschen Theorie’. *Sitzungsberichte der Königlich Preussischen Akademie der Wissenschaften*. 7: 189–196. Translated as, Antoci, S.; Loinger, A. (1999). ‘On the gravitational field of a mass point according to Einstein's theory’. https://arxiv.org/abs/physics/9905030

Ray J. Solomonoff. (1985). ‘The time scale of artificial intelligence: reflections on social effects’. *North-Holland Human Systems Management* 5:149–153.

Philip Trammell and Anton Korinek. (2023). ‘Economic Growth under Transformative AI,’ NBER Working Paper 31815, https://doi.org/10.3386/w31815.

Lizka Vaintrob and Owen Cotton-Barratt. (2025). ‘AI Tools for Existential Security’. https://www.forethought.org/research/ai-tools-for-existential-security

Eliezer Yudkowsky. (2001). ‘Creating friendly AI 1.0: The Analysis and Design of Benevolent Goal Architectures’, The Singularity Institute.